\def\year{2022}\relax
\documentclass[letterpaper]{article} 
\usepackage{mathtools}
\usepackage{aaai24}  
\usepackage{times}  
\usepackage{helvet}  
\usepackage{courier}  
\usepackage[hyphens]{url}  
\usepackage{graphicx} 
\usepackage{natbib}  
\usepackage{caption} 
\DeclareCaptionStyle{ruled}{labelfont=normalfont,labelsep=colon,strut=off} 
\usepackage{algorithm}
\usepackage{algorithmic}
\usepackage{latexsym}
\usepackage{amssymb}
\usepackage{amsmath}
\usepackage{mathtools}
\usepackage{amsthm}
\usepackage{amsmath}
\usepackage{amssymb}
\usepackage{bm}

\usepackage{newfloat}
\usepackage{listings}
\floatstyle{ruled}
\newfloat{listing}{tb}{lst}{}
\floatname{listing}{Listing}

\newtheorem{definition}{Definition}
\newtheorem{proposition}{Proposition}
\newenvironment{proofsketch}{%
  \proof}{\endproof}
\title{On Computing Plans with Uniform Action Costs}
\author{
Alberto Pozanco, Daniel Borrajo\footnote{On leave from Universidad Carlos III de Madrid}, Manuela Veloso
}
\affiliations{
J.P. Morgan AI Research\\
\{alberto.pozancolancho,daniel.borrajo,manuela.veloso\}@jpmorgan.com}

\usepackage{bibentry}

\begin{document}

\maketitle

\begin{abstract}
In many planning applications, agents might be interested in finding plans whose actions have costs that are as uniform as possible. 
Such plans provide agents with a sense of stability and predictability, which are key features when humans are the agents executing plans suggested by planning tools.
This paper adapts three uniformity metrics to automated planning, and introduces planning-based compilations that allow us to lexicographically optimize sum of action costs and action costs uniformity.
Experimental results in different benchmarks show that the reformulated tasks can be effectively solved in practice to generate uniform plans.
\end{abstract}

\section{Introduction}
Classical planning is the task of finding a plan, which is a sequence of deterministic 
actions such that, when applied in a given initial state, it results in a goal state~\cite{DBLP:books/daglib/0014222}.
Each action is associated with a non-negative cost, and the cost of a plan is defined as the sum of its actions' costs.
Plans with minimal cost are called optimal, and how to efficiently compute them accounts for a large part of automated planning research.

However, the real-world is full of applications where the sum of action costs is only one of the objectives that define an optimal plan~\cite{stewart1991multiobjective,geisser2022admissible,salzman2023heuristic}.
For example, in network routing one might be interested in finding widest shortest paths or shortest widest paths, in order to reduce bottlenecks and provide better service~\cite{sobrinho2001algebra}.
Or, in navigation domains, one can represent travel time as cost and be interested in finding shortest quickest or quickest shortest routes~\cite{golledge1995path}.

In this paper we introduce a novel objective that might be relevant in many planning applications: that of finding plans whose actions have costs that are as uniform as possible.
\begin{figure}
    \centering
    \includegraphics[scale=0.34]{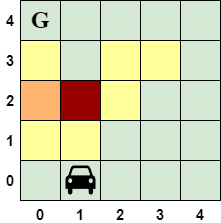}
    \caption{Navigation task where a driver wants to  reach location $G$. The color of each cell depicts the cost (travel time) of traversing the cell.}
    \label{fig:example}
\end{figure}
Consider the navigation scenario shown in Figure~\ref{fig:example}, where a car driver (located at position \texttt{1\_0}) is interested in finding a plan to reach location $G$ (located at \texttt{0\_4}).
To do so, the car can move in the four cardinal directions.
The color of each cell depicts the cost (travel time) of traversing the cell: visiting green, yellow, orange and red cells have a cost of $1$, $2$, $3$, and $4$ respectively.
In this case, there exist several cost-optimal plans, all having a cost of $9$, but requiring to traverse a different number of congested areas, with each of them having an associated congestion level.
Arguably, some drivers would prefer plans where they reduce the number or density of the traffic jams, thus being able to drive at the same pace.
Other drivers might even prefer longer travel times sacrificing cost-optimality in benefit of longer routes where they completely avoid traffic jams.
Moreover, not only drivers but also traffic authorities would prefer vehicles to keep a constant velocity, as this reduces congestions and $\mbox{CO}_2$ emissions~\cite{barth2009traffic}.

Financial planning~\cite{pozanco2023combining} is yet another example where computing plans with uniform actions is relevant.
In financial planning, a person wants to save some amount of money (e.g. $\$1,000$) in a given horizon (e.g. $4$ months).
The available actions are to save different quantities at each month, with increasing costs associated with higher savings levels due to the difficulty in saving more money.
People could prefer plans where they can save the same  money every month (e.g. $\$250$ each month) rather than plans where they should save very different amounts.
This preference can be justified by several reasons.
Firstly, a consistent savings plan provides individuals with a sense of stability and predictability. 
Knowing exactly how much they need to save each month allows them to budget and plan their expenses more effectively. 
This stability can alleviate financial stress and provide a sense of control over one's financial situation. 
Secondly, a fixed savings plan promotes discipline and consistency in saving habits. 
By committing to saving a specific amount each month, individuals are more likely to develop a routine and stick to their financial goals~\cite{hanna2010quantifying}. 

We are interested in lexicographically optimizing both objectives (in any order): sum of action costs and action costs uniformity.
There exist three main approaches to solve such multi-objective planning problems in the literature. 
The first one is using cost-algebraic A*~\cite{edelkamp2005cost}, which proves that A* returns an optimal solution in any \textit{cost algebra} (or their prioritized Cartesian product), not just in the traditional shortest path setting.
A cost measure is a cost algebra iff it is \textit{strictly isotone}, i.e., a solution that optimizes the cost measure is composed of optimal solutions to its subproblems.
Unfortunately, most uniformity measures do not satisfy this property, so we cannot use standard A*.
The second approach is to use multi-objective search algorithms such as NAMOA*~\cite{10.1145/1754399.1754400} or BOA*~\cite{ulloa2020simple,hernandez2023simple}.
These algorithms extend A* to compute the set of Pareto-optimal paths.
However, these algorithms are slower and consume more memory than standard A*, since they (i) need to explore the whole state space to return all the solutions in the Pareto front; and (ii) need to perform extra checks to prune dominated paths.
The third approach consists of reformulating the original planning task so that plans that optimally solve the new task are plans that lexicographically optimize the two objectives. 
This approach allows us to use any planner to solve such problems without the need of crafting domain-dependent heuristics for each objective.
For example, \citeauthor{katz2022producing}~(\citeyear{katz2022producing}) transform the action costs of the original planning tasks in order to generate shortest cost-optimal plans.
Here, we will reformulate the whole planning task, and not only its action costs, to generate uniform plans.

The main contributions of this paper are:
\begin{enumerate}
    \item Introduction of a novel bi-objective planning task with many real-world applications: finding cost-optimal plans whose actions have costs that are as uniform as possible.

    \item Adaptation of three dispersion metrics to automated planning.

    \item Three different compilations to produce plans that lexicographically optimize sum of action costs and the given dispersion metric.

    \item New benchmarks for multi-objective planning~\cite{salzman2023heuristic}.
\end{enumerate}

\section{Background}
We formally define a planning task as follows:
\begin{definition}
  A \textbf{planning task}
can be defined as a tuple $\Pi=\langle F,A,I,G\rangle $, where $F$ is a set of propositions, $A$ is a set of
instantiated actions, $I\subseteq F$ is an initial state, and $G\subseteq F$ is a goal state.  
\end{definition}

For each proposition $p \in F$ we introduce the term $\neg p$ representing the negation of $p$.
Propositions and their negations are called literals $L$.
A (partial) state $s$ is a consistent set of literals $l \subseteq L$, i.e., $s$ does not contain a proposition and its negation.
A complete state is a state which contains each proposition or its negation.
Each action $a\in A$ is
described by a set of preconditions (pre($a$)), which represent literals that must be true in a state to execute an
action, and a set of effects (eff($a$)), which are the literals that are added (add($a$) effects) or removed (del($a$)
effects) from the state after the action execution. The definition of each action might also include a cost $c(a) \in \mathbb{N}^0$ (the
default cost is one). 
We denote by $c(A)$, $\min(c(A))$ and $\max(c(A))$ the set of action costs, minimum action cost, and maximum action cost in the planning task, respectively.
The execution of an action $a$ in a state $s$ is defined by a function $\gamma$ such that $\gamma(s,a) = (s\setminus\mbox{del}(a))\cup\mbox{add}(a)$ iff $\mbox{pre}(a) \subseteq s$. 
The output of a planning task is a sequence of actions, called a plan, $\pi=(a_1,\ldots,a_n)$. The execution
of a plan $\pi$ in a state $s$ can be defined as:

\begin{small}
\[\Gamma(s,\pi)=\left\{\begin{array}{ll}
                        \Gamma(\gamma(s,a_1),(a_2,\ldots,a_n)) & \mbox{if } \pi\neq \emptyset\\
                        s & \mbox{if } \pi=\emptyset\\
                      \end{array}
                    \right.
\]
\end{small}

A state $s$ is reachable iff there exists a sequence of operators $\pi$ that when applied from $I$ reach $s$, i.e., $s \subseteq \Gamma(I,\pi)$.
With $S$ we refer to the set of all reachable states of the planning task.
A plan $\pi$ is valid if $G\subseteq\Gamma(I,\pi)$. The plan cost is commonly defined as
$c(\pi)=\sum_{a_i\in\pi} c(a_i)$, with $\bm{c}_\pi=(c(a_i), \ldots, c(a_n))$ denoting the action cost vector of a plan.
A plan with minimal cost is called optimal.
In the rest of the paper we assume plans are simple, i.e., they do not have loops visiting the same state more than once.
This is the case for any cost-optimal plan.

\section{Uniformest Cost-Optimal Planning Problem}
While classical planning focuses on finding cost-optimal plans, we seek plans that also optimize the uniformity of the plans' action costs.
Like \citeauthor{katz2022producing}~(\citeyear{katz2022producing}), we are not interested in any form of weighting both objectives, but in lexicographically optimizing them.

We seek to minimize two objectives , rather than minimizing plan's cost and maximizing its uniformity.
Therefore, we will minimize action costs dispersion in order to find uniform plans.
The \textit{dispersion of a plan}, denoted by $d(\pi)$, is defined as the variability of its action costs.
In other words, a plan only composed by actions of the same cost will have a lower dispersion, therefore being more uniform, than a plan composed by actions with different costs.
We denote by $\leq_d$ the partial order of plans defined by their action costs dispersion, i.e., $\pi \leq_d \pi^\prime$ iff $d(\pi) \leq d(\pi^\prime)$.
Likewise, we denote by $\leq_c$ the partial order of plans defined by their cost, i.e., $\pi \leq_c \pi^\prime$ iff $c(\pi) \leq c(\pi^\prime)$.
We will use $\preceq_{c,d}$ when we want to optimize cost, and break ties in favor of more uniform (less disperse) plans.
Formally, $\pi \preceq_{c,d} \pi^\prime$ if $c(\pi) < c(\pi^\prime)$ or if $c(\pi)=c(\pi^\prime)$ and $d(\pi) \leq d(\pi^\prime)$. 
The same notation and definition applies to the opposite case $\preceq_{d,c}$, where we want to optimize uniformity and break ties by cost.

Summarizing, for a given task $\Pi$ in standard classical planning we must identify a plan that is minimal among all plans with respect to $\leq_c$, or detect that the task is unsolvable.
On the other hand, in the uniformest cost-optimal planning problem, we must find a plan for $\Pi$ that is minimal among all plans with respect to $\preceq_{c,d}$ (or $\preceq_{d,c}$), or detect that the planning task is unsolvable.

\section{Dispersion Metrics}
Dispersion, also called variability, is a statistical term that refers to the spread between numbers in a data set.
In this case, we are interested in measuring the dispersion of vectors of plans action costs.
There exist many dispersion metrics: standard deviation, variance, range, or entropy, to name a few.
In this paper we will focus on the following three measures of statistical dispersion:

\begin{definition}
    \textbf{Number of different action costs} of a plan $\pi$: the cardinality of the set of plan's action costs.
    \begin{equation}
        \#(\pi) = |\{ \bm{c}_\pi \}|
    \end{equation}
\end{definition}

\begin{definition}
     \textbf{Delta} of a plan $\pi$: the largest difference between the cost of any two adjacent actions in the plan.
    \begin{equation}
        \Delta(\pi) = \max \Big(\bigcup_{i=1}^{|\pi| - 1}|c(a_i) - c(a_{i+1})|\Big)
    \end{equation}
\end{definition}

\begin{definition}
     \textbf{Range} of a plan $\pi$: the difference between the largest and smallest values in the costs of its actions.
    \begin{equation}
        R(\pi) = \max (\bm{c}_\pi) - \min (\bm{c}_\pi)
    \end{equation}
\end{definition}

The reason why we select the above dispersion metrics over others is two-folded.
First, we seek diverse set of dispersion metrics.
Well-known statistical measures such as standard deviation or variance provide the same information, while the three selected metrics have different semantics.
Range and Delta measure the difference between action costs in the plan, but while the former focuses on the entire plan, the latter only considers changes in two consecutive actions.
On the other hand, just measuring the number of different action costs offers less information, but might be easier to compute while still a good dispersion proxy.
We further investigate this in our evaluation.

Second, the above dispersion metrics have a nice property that other measures do not have: their potential values are bounded a priori by the available actions in the planning task.
While we can know all the possible values for Range, Delta and the number of different action costs by just inspecting the planning task, this is not possible for metrics such as standard deviation or entropy because the number of actions needed to achieve a goal is unbounded.
As we will discuss later, this is an important property that allows us to generate tractable reformulations in practice.

Let us exemplify these dispersion metrics and the plans we are interested in finding by using the navigation scenario introduced in Figure~\ref{fig:example}.
Table~\ref{tab:example_metrics} shows four possible plans to reach the goal in this scenario.
\begin{table}[]
    \centering
    \small
    \begin{tabular}{l|c|c|c|c}
        \multicolumn{1}{c|}{$\pi$} & $\#(\pi)$ & $\Delta(\pi)$ & $R(\pi)$ & $c(\pi)$ \\ \hline
         \includegraphics[scale=0.4]{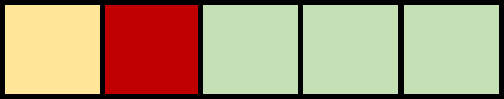}& 3 & 3 & 3 & 9 \\ \hline
        \includegraphics[scale=0.4]{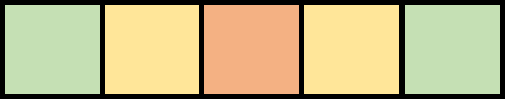} & 3 & 1 & 2 & 9 \\ \hline
        \includegraphics[scale=0.4]{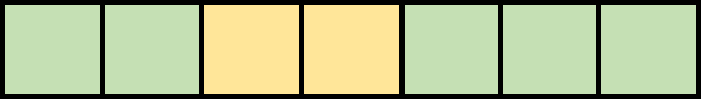} & 2 & 1 & 1 & 9 \\ \hline
        \includegraphics[scale=0.4]{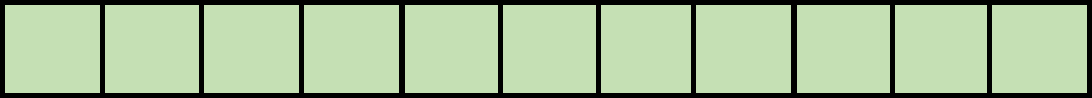} & 1 & 0 & 0 & 11 \\ \hline      
    \end{tabular}
    \caption{Four alternative plans to reach $G$ in the scenario shown in Figure~\ref{fig:example}. Plans are depicted by sequences of squares (from left to right) that show the driver's path from $I$ to $G$. Columns show dispersion and cost measures and cells in the table indicate the metric obtained by each plan.}
    \label{tab:example_metrics}
\end{table}
Each plan is depicted by a sequence of squares (from left to right) that show the path of the driver from its initial state to the goal.
Columns in this table show the three dispersion metrics, i.e., Range, Delta, and Number of different action costs, as well as the cost of the plan.
The plans in the first three rows are cost-optimal and have an associated cost (travel time) of $9$.
However, these plans are very different in terms of action cost dispersion.
The first plan combines driving through extremely (red) and slightly (green) congested areas, yielding the higher (worse) values for all the dispersion metrics.
This plan will not be to the liking of drivers who prefer to maintain a steady pace.
The second plan has a lower Delta and Range, since the change in congestion levels occurs more progressively and extremely congested areas are not traversed.
The third plan has the lowest values for all the dispersion metrics along the cost-optimal plans, and would be the one selected by drivers that prefer more uniform driving experience while not sacrificing cost (driving time).
Finally, the fourth plan might be more appealing to drivers that might be willing to sacrifice cost-optimality in benefit of more uniform routes where they can completely avoid traffic jams.
Among all the plans that minimize dispersion, this is the one with the lowest cost.

\section{Computing Plans with Uniform Action Costs}

We present three different compilations, one for each of the dispersion metrics previously described.
Although some of the compilations could be represented more naturally using conditional effects or numerical planning, we opted for classical planning compilations, as the resulting tasks can be efficiently solved optimally by a wider variety of planners.
In particular, unlike other compilations that take as input the lifted planning task (represented in PDDL~\cite{mcdermott1998pddl}) and return a new lifted planning task~\cite{DBLP:conf/aips/Gragera0O023}, our reformulation operates at the grounded (\textsc{strips}) level.
We opted for this approach as in some planning tasks the cost of the actions is not known at the lifted level, but only after the task has been grounded.

The intuition behind all the compilations is to extend a standard planning task with new propositions that keep track of the action cost distribution.
They also extend the original planning task with actions that either (i) update these propositions; or (ii) increase the total cost depending on the value of the dispersion metric induced by the plan.
The cost of these actions will depend on the objective function that we try to lexicographically optimize.
We will use $\omega_d \in \mathbb{N}$ to weigh the importance of getting uniform plans.

Next, we introduce the compilations in increasing order of complexity, according to the number of extra propositions and actions they require.

\subsection{Compilation 1: Number of Different Action Costs}
This is the most naive dispersion metric, as it only considers the cardinality of the plan's action cost set.
It also requires the least complex compilation, as this metric is strictly isotone~\cite{edelkamp2005cost}.
Given a planning task $\Pi$, we extend $F$ with a set of propositions $F_u = \bigcup_{i \in  c(A)} \{ \mbox{used}_i \}$, which keep track of the costs that have been used by a plan.
We also update the set of original actions $A$ to associate each action $a \in A$ a different cost depending on whether that action cost has already been used in the plan or not.
We create two new actions for each original action $a \in A$, effectively dividing actions into two sets, $A_u$ and $A_{\neg u}$.
\begin{itemize}
    \item $A_{u}$, which represents the original actions that can be executed when their cost is already present in the plan. Each action $a_{u} \in A_{u}$ is defined as follows:
    \begin{itemize}
    \small
        \item $\mbox{pre}(a_u) = \mbox{pre}(a) \cup \{ \mbox{used}_{c(a)} \}$

        \item $\mbox{eff}(a_u) = \mbox{eff}(a)$

        \item $c(a_u) = c(a)$
    \end{itemize}

    \item $A_{\neg u}$, represent the original actions that can be executed when their cost is not present in the plan yet. Each action $a_{\neg u} \in A_{\neg u}$ is defined as follows:
    \begin{itemize}
    \small
        \item $\mbox{pre}(a_{\neg u}) = \mbox{pre}(a) \cup \{ \neg\mbox{used}_{c(a)} \}$

        \item $\mbox{eff}(a_{\neg u}) = \mbox{eff}(a) \cup \{ \mbox{used}_{c(a)} \}$

        \item $c(a_{\neg u}) = c(a) + \omega_d$
    \end{itemize}
\end{itemize}

The difference between both sets of actions is that the ones in $A_{\neg u}$ set the unseen cost $c(a)$ to used, and increase the total cost with the weight $\omega_d$, thus effectively penalizing the use of actions with different action costs to achieve the goal.

The new planning task is defined as follows:
\begin{definition}
    Given an original planning task $\Pi$, a \textbf{compiled task} $\Pi_{\#}$ that lexicographically \textbf{optimizes sum of action costs and number of different action costs} as dispersion metric is a tuple
    \begin{equation*}
        \Pi_{\#} = \langle F^\prime, A^\prime, I^\prime, G, \omega_d \rangle
    \end{equation*}
    where,
    \begin{itemize}
    \small
        \item $F^\prime = F \cup F_u$
        
        \item $A^\prime = A_u \cup A_{\neg u}$

        \item $I^\prime = I \cup \bigcup_{i \in c(A) } \{ \neg \mbox{used}_i\}$
    \end{itemize}
\end{definition}

This compilation grows polynomially with $|A|$ and generates $|A^\prime| = 2 \times |A|$ actions, as we need to have two versions of each action, depending on whether its cost has already been used in the plan or not.
\begin{proposition}
    If $\Pi$ is solvable, $\Pi_\#$ is also solvable.
\end{proposition}
\begin{proofsketch}
    Let $\pi$ be a solution for $\Pi$.
    It is trivial to see that there exists a plan $\pi^\prime$ that solves $\Pi_\#$ comprised by the same sequence of actions as $\pi$, only varying their version, i.e., if their cost has already been used in the plan ($A_u$) or not ($A_{\neg u}$).
    The applicability of these actions is equivalent to their $A$ counterparts, with the extra precondition of the \textit{used} proposition of that cost, which appear in $I^\prime$ and are updated by the actions in $A^\prime$.  
\end{proofsketch}

Next we prove that $\preceq_{c,d}$ (or $\preceq_{d,c}$) is preserved in the reformulated task.
The proof is similar to that of~\cite{katz2022producing}, where the authors prove their reformulated task can generate shortest cost-optimal plans.
The main idea is to set $\omega_d$ so that for any plan, the extra cost incurred by introducing new action costs does not surpass the sum of action costs (or the opposite in case $\preceq_{d,c}$).

\begin{proposition}
    Let $\pi$ and $\pi^\prime$ be any two simple plans for $\Pi$, and $\pi_\#$ and $\pi^\prime_\#$ their counterparts in $\Pi_\#$.
    There exists a value for $\omega_d$ such that $\pi \preceq_{c,d} \pi^\prime$ iff $\pi_\# \leq_c \pi^\prime_\#$.   
\end{proposition}
\begin{proofsketch}
    Since the plans are simple, the cost of any $\pi$ that solves $\Pi$ is bounded by $|S| \times \max(c(A))$.
    Let $C^->0$ be smaller or equal to the smallest non-zero cost difference $|c(\pi) - c(\pi^\prime)|$ between two plans that solve $\Pi$.
    Let $D^+$ be larger than the max different action costs difference $|\#(\pi) - \#(\pi^\prime)|$ between two plans that solve $\Pi$.
    We define $\omega_d \coloneq \frac{C^-}{D^+} > 0$.

    We distinguish two cases.
    If $c(\pi) = c(\pi^\prime)$, then $\pi \preceq_{c,d} \pi^\prime$ iff $\#(\pi) \leq \#(\pi^\prime)$.
    By our compilation we have that 
    \begin{equation}
        c(\pi_\#) = c(\pi) + \omega_d \#(\pi)  \leq c(\pi^\prime) + \omega_d\#(\pi^\prime)  = c(\pi^\prime_\#)
        \label{eq}
    \end{equation}
    so when $c(\pi) = c(\pi^\prime)$ this can be simplified to $\#(\pi) \leq \#(\pi^\prime)$, iff $\pi_\# \leq_c \pi^\prime_\#$. 
    
    On the other hand, if $\pi \preceq_{c,d} \pi^\prime$ and $c(\pi) \neq c(\pi^\prime)$ then $c(\pi) < c(\pi^\prime)$.
    Since $c(\pi^\prime) - c(\pi)$ is non-zero, it must be $\geq C^-$.
    By the choice of $D^+$, $\#(\pi) - \#(\pi^\prime) < D^+$.
    We get that $c(\pi^\prime_\#) - c(\pi_\#) = c(\pi^\prime) + \omega_d \#(\pi^\prime)  - c(\pi) - \omega_d\#(\pi) = c(\pi^\prime) - c(\pi) - \omega_d(\#(\pi^\prime)-\#(\pi)) > c(\pi^\prime) - c(\pi) - \omega_d D^+ = c(\pi^\prime) - c(\pi) - C^- \geq 0$.
    Overall we get that $c(\pi^\prime_\#) > c(\pi_\#)$, thus $\pi_\# \leq_c \pi^\prime_\#$.
    In the opposite direction, if $\pi_\# \leq_c \pi^\prime_\#$, then $c(\pi_\#) \leq c(\pi^\prime_\#)$.
    By (\ref{eq}), $c(\pi) + \omega_d \#(\pi) \leq c(\pi^\prime) + \omega_d \#(\pi^\prime)$.
    Since $\#(\pi) - \#(\pi^\prime) < D^+$, it holds that $\omega_d(\#(\pi^\prime) - \#(\pi)) < \omega_d D^+ = C^-$.
    By the definition of $C^-$, $C^- > c(\pi) - c(\pi^\prime)$.
    This implies $c(\pi) < c(\pi^\prime)$ and thus $\pi \preceq_{c,d} \pi^\prime$.
\end{proofsketch}

As in the case of~\cite{katz2022producing}, since we only focus on integer action costs ($C^- = 1$), it is sufficient for the proof if $|\#(\pi_\#) - \#(\pi^\prime_\#)| < 1/\omega_d$.
This guarantees that if we set $\omega_d < 1 / D^+$, $\preceq_{c,d}$ is preserved.

Analogously it is easy to see that if we set $\omega_d > C^+$, where $C^+$ is the largest cost difference between two plans ($|S| \times \max(c(A))$), then $\pi \preceq_{d,c} \pi^\prime$ iff $c(\pi_\#) \leq c(\pi^\prime_\#)$.

\subsection{Compilation 2: Delta}
Delta measures the largest jump in action costs in two consecutive actions in a plan.
Minimizing it can be useful for agents content with plans having disperse action costs, as long as the cost changes are gradual rather than sudden.
This dispersion metric is not strictly isotone, so the compilation to optimize it becomes more complex.

In this compilation we will use $abs(c(A))$ (or just $abs$) to refer to the set of absolute values obtained after subtracting all the cost pairs in the domain, i.e., all the possible Delta values.
These values are given by the following formula:
\begin{equation}
    abs(c(A)) = \bigcup_{m,n \in c(A) \cup \{ 0\}} \{ |m-n|\}
\end{equation}
We cannot define a general formula based on the size of $c(A)$ to compute the size of $abs$, as it does not solely depend on the number of possible costs, but also on the arithmetic relationship of its elements.
For example, $c(A) = \{ 0, 5, 10 \}$ yields a set of possible Delta values $abs = \{ 0,5, 10\}$, while $c(A) = \{ 1, 5, 10 \}$ yields $abs = \{ 0,1,4,5,9, 10\}$.
However, we can upper bound the size of this set by the binomial coefficient ${|c(A)| + 1 \choose 2}$. In the worst case, all the pairs of the set will yield a different absolute value, plus 0, since that is the result of the subtraction of any pair $(x,x)$.

Given a planning task $\Pi$, we extend $F$ with the following sets of propositions:
\begin{itemize}
    \item $F_{prev} = \bigcup_{i \in c(A)} \{ \mbox{prev\_cost}_i\}$, which is a set of propositions that keep track of the last executed action's cost.

    \item $F_{\delta} = \bigcup_{i \in abs} \{ \mbox{delta}_i\}$, which is a set of propositions that keep track of the Delta between the current and the previous actions executed in the plan.

    \item $F_{\Delta} = \bigcup_{i \in abs} \{ \mbox{max\_delta}_i\}$, which is a set of propositions that track the largest Delta in the  plan.

    \item $check$, which is a proposition used as a flag to indicate when the max\_delta propositions should be checked and updated.

    \item $end$, which is a proposition that represents the fact that all goals have been achieved.
\end{itemize}

We also extend $A$ with three types of actions. 
The first type $A_{\delta}$ updates the original actions to keep track of the prev\_cost counter and the current value of Delta between the current and the previous action.
Each action $a_{\delta} \in A_{\delta}$ is defined as follows:
\begin{itemize}
\small
\setlength{\itemindent}{0.3cm}
    \item[\textbf{--}] $\mbox{pre}(a_{\delta}) = \mbox{pre}(a) \cup \{ \mbox{prev\_cost}_i, \neg check\}$

    \item[\textbf{--}] 
    \begin{tabbing}
    \ \ \ $\mbox{eff}(a_{\delta})$ =  \= $\mbox{eff}(a) \cup$ \\
    \ \ \ \ \ \ \ \ \ \ \ \ \ \ $\{ \neg \mbox{prev\_cost}_i, \mbox{prev\_cost}_{c(a)}, \mbox{delta}_{|i-c(a)|}, check \}$
    \end{tabbing}

    \item[\textbf{--}] $c(a_{prev}) = c(a)$
\end{itemize}

The second type of actions will appear interleaved in the plan between the updated original actions. For each pair of costs $i, j \in c(A)$, if $i>j$, we generate a new action in a new set $A_{\Delta}$. Otherwise, we generate a new action in another new set $A_{\neg\Delta}$. 
Since we are operating over the grounded task, we can make the check when generating the compilation.
\begin{itemize}
    \item $A_{\Delta}$, which are new actions that update the counter when delta is larger than max\_delta. Each action $a_{\Delta} \in A_{\Delta}$ is defined as follows:
    \begin{itemize}
    \small
        \item $\mbox{pre}(a_\Delta) = \{ check, \mbox{delta}_i, \mbox{max\_delta}_j\}$

        \item $\mbox{eff}(a_\Delta) = \{ \neg check, \neg \mbox{delta}_i, \neg \mbox{max\_delta}_j, \mbox{max\_delta}_i \}$

        \item $c(a_\Delta) = 0$
    \end{itemize}

    \item $A_{\neg\Delta}$, which are new actions that do not update the counter when delta is lower or equal than max\_delta, just turn off the $check$ flag, so $A_\delta$ actions can be executed again. Each action $a_{\neg\Delta} \in A_{\neg\Delta}$ is defined as follows:
    \begin{itemize}
    \small
        \item $\mbox{pre}(a_{\neg\Delta}) = \{ check, \mbox{delta}_i, \mbox{max\_delta}_j \}$

        \item $\mbox{eff}(a_{\neg\Delta}) = \{ \neg check, \neg \mbox{delta}_i \}$

        \item $c(a_{\neg\Delta}) = 0$
    \end{itemize}
\end{itemize}

Finally, the third type of actions can only be executed at the end of planning, i.e., once a goal state has been reached. They increase the total cost of the plan depending on its Delta (Definition 3), i.e., the value of the $\mbox{max\_delta}$ proposition.
We refer to this set of actions as $A_{end}$, and each action $a_{end} \in A_{end}$ is defined as follows:
\begin{itemize}
\small
\setlength{\itemindent}{0.3cm}
    \item[\textbf{--}] $\mbox{pre}(a_{end}) = G \cup \{ \neg check , \mbox{max\_delta}_i\}$

    \item[\textbf{--}] $\mbox{eff}(a_{end}) = \{ end \}$

    \item[\textbf{--}] $c(a_{end}) = i \times \omega_d$
\end{itemize}

The new planning task is defined as follows:
\begin{definition}
    Given an original planning task $\Pi$, a \textbf{compiled task} $\Pi_\Delta$ that lexicographically \textbf{optimizes sum of action costs and Delta as dispersion metric} is a tuple
    \begin{equation*}
        \Pi_\Delta = \langle F^\prime, A^\prime, I^\prime, G^\prime, \omega_d \rangle
    \end{equation*}
    where,
    \begin{itemize}
    \small
        \item $F^\prime=F\cup F_{prev} \cup F_{\delta} \cup F_{\Delta} \cup \{ check, end \}$

        \item $A^\prime=A_\delta \cup A_{\Delta} \cup A_{\neg\Delta} \cup A_{end}$

        \item $I^\prime=I \cup \{ \mbox{prev\_cost}_{\min(c(A))},\mbox{max\_delta}_{\min(c(A))}\}$

        \item $G^\prime=\{end\}$
    \end{itemize}
\end{definition}

This compilation generates a polynomial number of actions that is given by the following formula:
\begin{equation*}
    |A^\prime| = \Big(\overbrace{|A| \times |c(A)|}^{A_{\delta}}\Big) + \overbrace{|abs|^2}^{A_{\Delta},A_{\neg\Delta}} + \overbrace{|abs|}^{A_{end}}
\end{equation*}
The first part of the formula refers to the different possible values of $\mbox{prev\_cost}$ appearing in the precondition of the $A_{\delta}$ actions.
The second part refers to the actions that update the max\_delta propositions, i.e., one action for each combination of $\mbox{delta}_i, \mbox{max\_delta}_j, \forall_{i,j \in abs}$.
The third part refers to the potential values of max\_delta at the end of planning.
Following a similar reasoning as in Propositions 1 and 2, it is easy to see that this compilation is also complete, sound, and can preserve $\preceq_{c,d}$ or $\preceq_{d,c}$ if we set $\omega_d$ as a relation between the max/min Delta and cost differences between two plans.

\subsection{Compilation 3: Range}
Range is one of the most widely employed dispersion metrics, since it is easy to compute and is known to correctly approximate standard deviation $\sigma$ under different distributions by applying the formula $\sigma\approx\frac{\mbox{Range}}{4}$~\cite{shiffler1980upper}.
This dispersion metric is again not strictly isotone.

Given a planning task $\Pi$, we extend $F$ as follows:
\begin{itemize}
    \item $F_{\min} = \bigcup_{i \in  c(A)} \{ \mbox{min\_cost}_i \}$, which is a set of propositions that keep track of the cost of the least costly action executed in a plan.

    \item $F_{\max} = \bigcup_{i \in  c(A)} \{ \mbox{max\_cost}_i \}$, which is a set of propositions that keep track of the cost of the most costly action executed in a plan.

     \item $end$, which is a proposition that represents the fact that all goals have been achieved.
\end{itemize}

We also extend the original set of actions with two types of actions.
The first type updates the original actions to keep track of the min and max counters.
We divide them into the following subsets of actions, which cover all possible ways in which the counters can be updated.
Like in the Delta compilation, we can do this because we know all the action costs combinations beforehand.
\begin{itemize}
    \item $A_{both}$, which are the original actions that update both counters when the cost of the action is (i) lower than the minimum, and (ii) greater than the maximum costs employed in the plan. Each action $a_{both} \in A_{both}$ is defined as follows:
    \begin{itemize}
    \small
        \item $\mbox{pre}(a_{both}) = \mbox{pre}(a) \cup \{ \mbox{min\_cost}_i, \mbox{max\_cost}_j\}$

        \item $\mbox{eff}(a_{both}) = \mbox{eff}(a) \cup$\begin{multline*}
            \{ \neg \mbox{min\_cost}_i, \mbox{min\_cost}_{c(a)}, \neg \mbox{max\_cost}_j, \mbox{max\_cost}_{c(a)}\}
        \end{multline*}

        \item $c(a_{both}) = c(a) $
    \end{itemize}

    \item $A_{\min}$, which are the original actions that update the $\mbox{min\_cost}$ counter when the cost of the action is (i) lower than the minimum cost, and (ii) lower or equal than the maximum cost employed in the plan. Each action $a_{\min} \in A_{\min}$ is defined as follows:
    \begin{itemize}
    \small
        \item $\mbox{pre}(a_{\min}) = \mbox{pre}(a) \cup \{ \mbox{min\_cost}_i, \mbox{max\_cost}_j\}$

        \item $\mbox{eff}(a_{\min}) = \mbox{eff}(a) \cup
            \{ \neg \mbox{min\_cost}_i, \mbox{min\_cost}_{c(a)}\}$

        \item $c(a_{\min}) = c(a)$
    \end{itemize}

    \item $A_{\max}$, which are the original actions that update the $\mbox{max\_cost}$ counter when the cost of the action is (i) greater or equal than the minimum cost, and (ii) greater than the maximum cost employed in the plan. Each action $a_{\max} \in A_{\max}$ is defined as follows:
    \begin{itemize}
    \small
        \item $\mbox{pre}(a_{\max}) = \mbox{pre}(a) \cup \{ \mbox{min\_cost}_i, \mbox{max\_cost}_j\}$

        \item $\mbox{eff}(a_{\max}) = \mbox{eff}(a) \cup
            \{ \neg \mbox{max\_cost}_j, \mbox{max\_cost}_{c(a)}\}$

        \item $c(a_{\max}) = c(a)$
    \end{itemize}

    \item $A_{none}$, which are the original actions that do not update any counter when the cost of the action is (i) greater or equal than the minimum cost, and (ii) lower or equal than the maximum cost employed in the plan. Each action $a_{none} \in A_{none}$ is defined as follows:
    \begin{itemize}
    \small
        \item $\mbox{pre}(a_{none}) = \mbox{pre}(a) \cup \{ \mbox{min\_cost}_i, \mbox{max\_cost}_j\}$

        \item $\mbox{eff}(a_{none}) = \mbox{eff}(a)$

        \item $c(a_{none}) = c(a)$
    \end{itemize}
\end{itemize}

The second type of actions can only be executed at the end of planning, i.e., once a goal state has been reached, and increase the total cost of the plan depending on its Range.
We create one action for each possible value of $abs$ (Equation 4).
We refer to this set of actions as $A_{end}$, and each action $a_{end} \in A_{end}$ is defined as follows:
\begin{itemize}
\small
    \setlength{\itemindent}{0.3cm}
    \item[\textbf{--}] $\mbox{pre}(a_{end}) = \mbox{G} \cup \{ \mbox{min\_cost}_i , \mbox{max\_cost}_j\}$
    \item[\textbf{--}] $\mbox{eff}(a_{end}) = \{ end \}$
    \item[\textbf{--}] $c(a_{end}) = (j-i) \times \omega_d$
\end{itemize}

The new planning task is defined as follows:
\begin{definition}
    Given an original planning task $\Pi$, a \textbf{compiled task} $\Pi_R$ that lexicographically \textbf{optimizes sum of action costs and Range as dispersion metric} is a tuple
    \begin{equation*}
        \Pi_R = \langle F^\prime, A^\prime, I^\prime, G^\prime, \omega_d \rangle
    \end{equation*}
    where,
    \begin{itemize}
    \small
        \item $F^\prime = F \cup F_{\min} \cup F_{\max} \cup \{ end \}$

        \item $A^\prime = A_{both} \cup A_{\min} \cup A_{\max} \cup A_{none} \cup A_{end}$

        \item $I^\prime = I \cup \{ \mbox{min\_cost}_{\max(c(A))}, \mbox{max\_cost}_{\min(c(A))} \}$

        \item $G^\prime = \{ end \}$
    \end{itemize}
\end{definition}

This compilation generates a polynomial number of actions that is given by the following formula:
\begin{equation*}
    |A^\prime| = \overbrace{(|A| \times |c(A)|^2)}^{A_{both}, A_{\min}, A_{\max}, A_{none}} + \overbrace{|c(A)|^2}^{A_{end}}
\end{equation*}
The first part of the formula refers to the different variations of the original actions that we create for each possible value of the min\_cost and max\_cost counters.
The second part refers to the actions that achieve the $end$ goal proposition.
The number of these actions is given by all the combinations of the min and max counters values.
Following a similar reasoning as in Propositions 1 and 2, it is easy to see that this compilation is also complete, sound, and can preserve $\preceq_{c,d}$ or $\preceq_{d,c}$ if we set $\omega_d$ as a relation between the max/min Range and cost differences between two plans.

\section{Evaluation}
\subsection{Experimental Setting}

\subsubsection{Benchmark.} To conduct the evaluation, we need planning tasks with at least two different action costs.
We analyzed the tasks in the Planning Domains repository\footnote{\url{https://github.com/AI-Planning/classical-domains}}, and picked those domains without conditional effects that have problems that, after grounding,  satisfy $|c(A)| \geq 2$.
In case some domains were duplicated (either from different years or tracks), we selected the problems of the latest optimal track.
Table~\ref{tab:benchmarks} summarizes our benchmark, which consists of $387$ problems divided into $19$ domains.
The first column indicates the domain, as well as the maximum number of different action costs in any problem within the domain ($\max |c(A)|$).
The second column represents the distribution of costs in all the problems. 
Larger points along the segment indicate the given cost is present in more problems.
The last two rows, \textsc{finance} and \textsc{navigation} are the two novel domains we discussed in the introduction for which we generated the corresponding domain and randomly generated problems sets.
They are thought to be challenging for uniformest cost-optimal planning, as their problems have many more different action costs than standard planning tasks. 
\begin{table}
\footnotesize
    \centering
    \begin{tabular}{r|c}
    \textbf{Domain ($\max|c(A)|$)} & \textbf{Actions Costs Distribution} \\ \hline 
       scanalyzer (2)  & \includegraphics[scale=0.12]{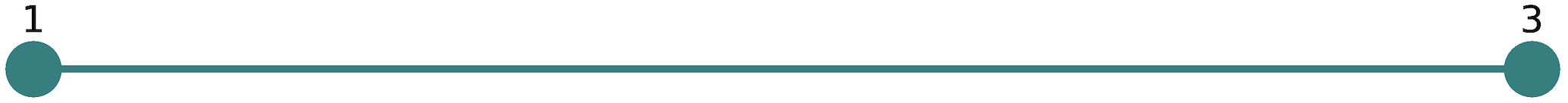}\\ 
        sokoban (2) & \includegraphics[scale=0.12]{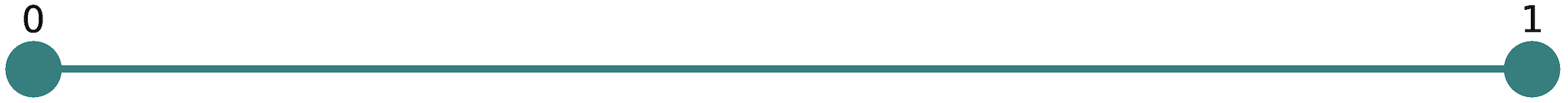}\\
        pegsol (2) & \includegraphics[scale=0.12]{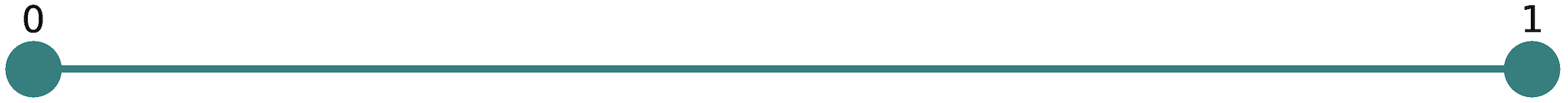}\\ 
        openstacks (2) & \includegraphics[scale=0.12]{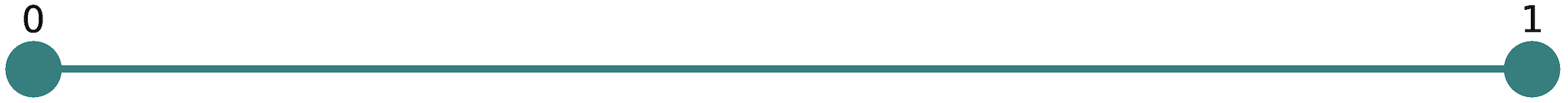}\\
        barman (2) & \includegraphics[scale=0.12]{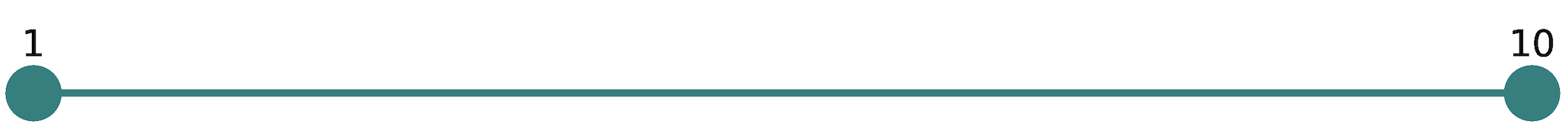}\\ 
        petri-net (2) & \includegraphics[scale=0.12]{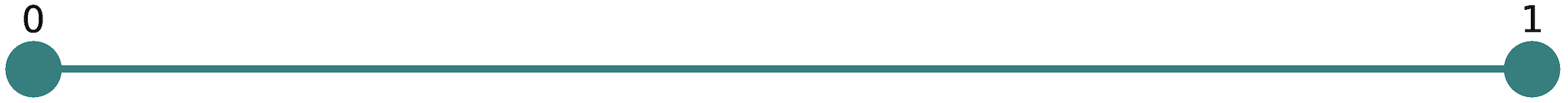}\\
        tetris (3) & \includegraphics[scale=0.12]{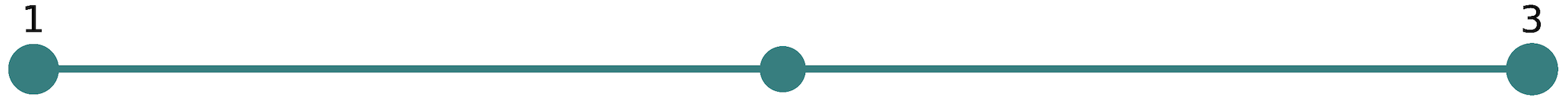}\\
        gedit (3) & \includegraphics[scale=0.12]{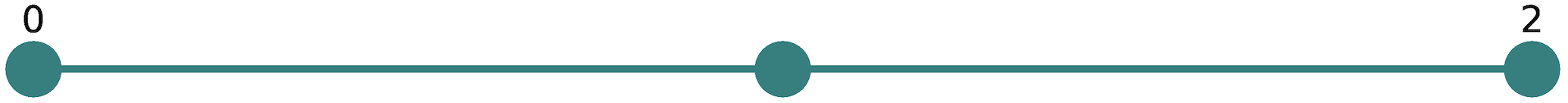}\\
        floortile (4) & \includegraphics[scale=0.12]{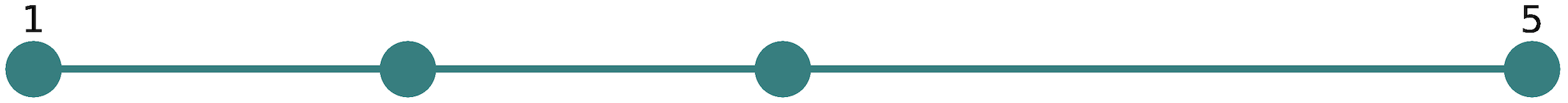}\\
        woodworking (6) & \includegraphics[scale=0.12]{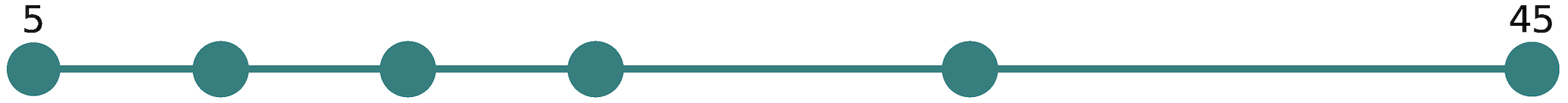}\\
        agricola (6) & \includegraphics[scale=0.12]{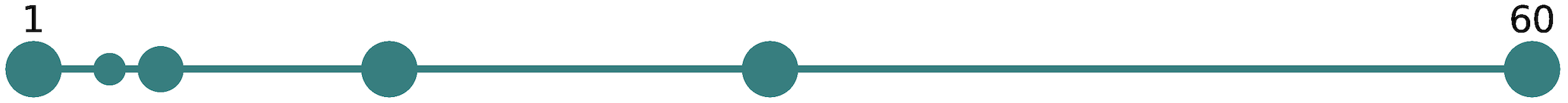}\\
        elevators (10) & \includegraphics[scale=0.12]{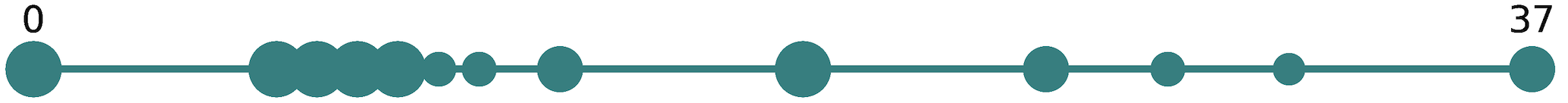}\\ 
        cybersec (11) & \includegraphics[scale=0.12]{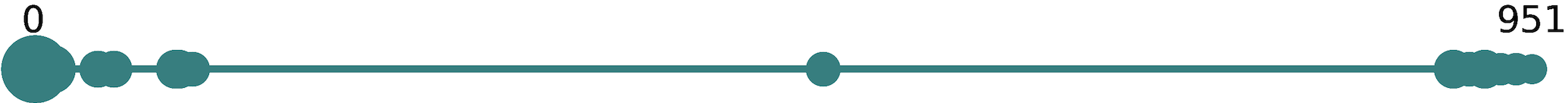}\\ 
        organic-synthesis (19) & \includegraphics[scale=0.12]{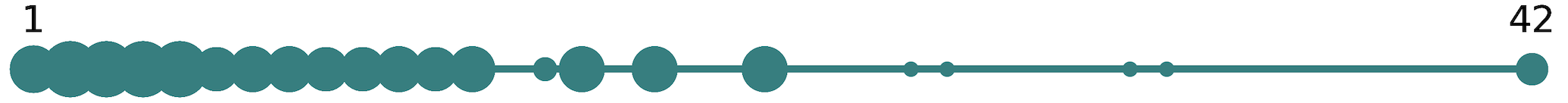}\\ 
        parcprinter (22) & \includegraphics[scale=0.12]{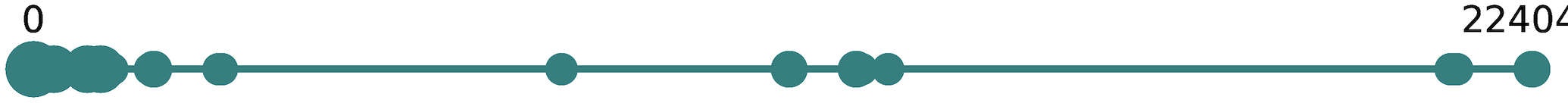}\\ 
        transport (27) & \includegraphics[scale=0.12]{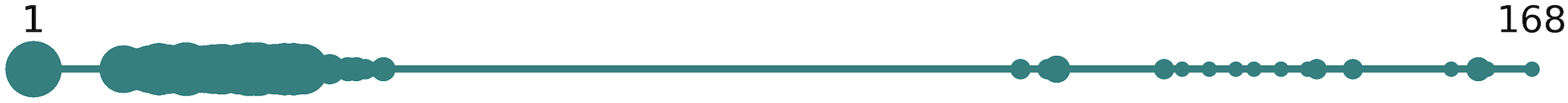}\\
        data-network (33) & \includegraphics[scale=0.12]{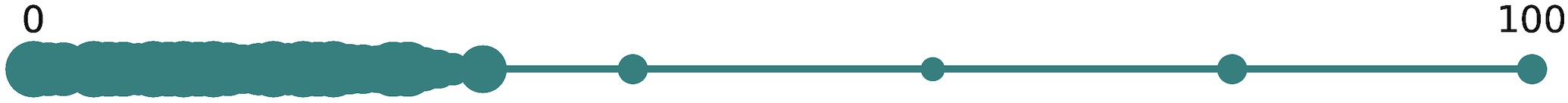}\\
        finance (41) & \includegraphics[scale=0.12]{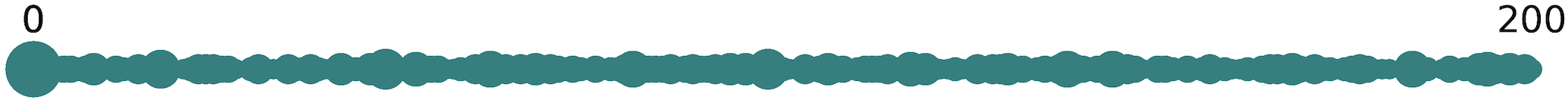}\\
        navigation (101) & \includegraphics[scale=0.12]{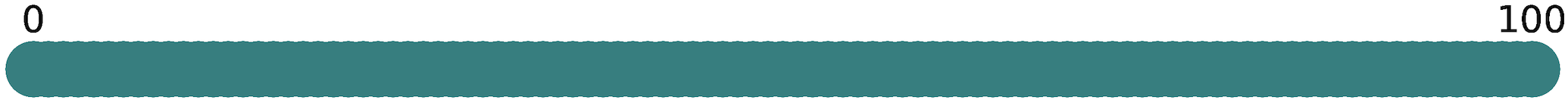}\\
    \end{tabular}
    \caption{Actions costs distribution across our benchmark.}
    \label{tab:benchmarks}
\end{table}

\subsubsection{Approaches.}
We evaluate the three compilations, namely $\Pi_{\#}, \Pi_{\Delta}$ and $\Pi_R$, on the above benchmark.
Since most planners only accept integer costs, we perform a similar cost transformation as in~\cite{katz2022producing} and set the weights as follows.
When solving $\preceq_{c,d}$, we set $c(a) = c(a) \times 10^6$ and $\omega_d=1$.
Since $10^6$ is a loose upper bound on the maximum dispersion a plan can have in our benchmark, we keep the theoretical properties.
When solving $\preceq_{d,c}$, we set $\omega_d=2\times10^6$, which is again a loose upper bound on the cost of the simple plans in our benchmark.
Given a compilation $C$, we will refer to these configurations as $\Pi_C^{c,d}$ or $\Pi_C^{d,c}$, respectively.
We compare these $3\times2=6$ compilations against two \textit{baselines}.
The first one uses a standard planner solving task $\Pi$.
We use this to understand the overhead incurred by the compilations compared to solving the simpler task where only cost is optimized.
The second one computes all the optimal plans of the standard planning task, which we refer to as ${\cal P}(\Pi)$.
We use this to understand if solving the compiled planning tasks is faster than computing ${\cal P}(\Pi)$ and then post-processing that set to find the most uniform plan.
These two baselines can only be used to compare against the $\Pi^{c,d}$ compilations, since the $\Pi^{d,c}$ compilations are solving a different task.
However, we put them together so we can understand (i) which task is easier; and (ii) how far solving $\Pi^{c,d}$ is from the optimal dispersion achieved by solving $\Pi^{d,c}$.

\subsubsection{Reproducibility.} We solve all the planning tasks ($\Pi, \Pi_{\#}, \Pi_{\Delta}$ and $\Pi_R$) and their different variations using the \textsc{seq-opt-lmcut} configuration of Fast Downward~\cite{DBLP:journals/jair/Helmert06}, which runs $A^*$ with the admissible \textsc{lmcut} heuristic to compute an optimal plan. 
For computing the set of optimal plans ${\cal P}(\Pi)$, we use \textsc{ForbidIterative}~\cite{katz2018novel} with a limit of $100,000$ plans to avoid disk overflows.
We chose this planner over other alternatives such as \textsc{Sym-K}~\cite{speck2020symbolic} because its search process is more similar to that of \textsc{seq-opt-lmcut}, and therefore comparing their results is more fair.
Experiments were run on an Intel Xeon E5-2666 v3 CPU @ 2.90GHz x 8 processors with a 8GB memory bound and a time limit of 1800s.
Code and benchmarks are available upon request.

\subsection{Results}

\subsubsection{Coverage and Execution Time Overhead.}
First, we want to understand how hard solving our compiled tasks is.
Table~\ref{tab:coverage} shows the coverage of each compilation and the two baselines across all domains and problems.
\begin{table}[]
\setlength{\tabcolsep}{3pt}
    \centering
    \small
    \begin{tabular}{c|c|c|c|c|c|c|c|c}
         & $\Pi$ & $\Pi_{\#}^{d,c}$ & $\Pi_{\#}^{c,d}$ & $\Pi_{\Delta}^{d,c}$ & $\Pi_{\Delta}^{c,d}$ & $\Pi_{R}^{d,c}$ & $\Pi_{R}^{c,d}$ & ${\cal P}(\Pi)$ \\ \hline
        Coverage & $225$ & $167$ & $204$ & $100$ & $115$ & $105$ & $125$ & $115$ 
    \end{tabular}
    \caption{Number of problems solved by each compilation.}
    \label{tab:coverage}
\end{table}
As expected, solving the standard planning task where only cost is optimized is easier, and the planner can find a cost-optimal plan for $225$ instances.
Optimizing for cost and breaking ties in favor of less disperse plans ($\Pi^{c,d}$ compilations) tends to be easier than doing the opposite, regardless of the dispersion metric.
Taking a look to the specific metrics, minimizing the number of different action costs ($\Pi_{\#}$ compilations) is somewhat comparable to standard planning, and definitely more feasible than computing all the cost-optimal plans.
This is not the case when jointly optimizing cost and Delta or Range, where we get slightly lower coverages than when solving ${\cal P}(\Pi)$. 
This difference in coverage comes mainly from domains such as \textsc{finance} or \textsc{navigation}, which are difficult for uniformest cost-optimal planning due to their large number of different action costs, but easier for top-quality planning, as they contain few cost-optimal plans.
However, by post-processing all the cost-optimal plans we would only get those that optimize these metrics as a second objective, i.e., plans generated by the $\Pi^{c,d}$ compilations, and not those plans that prioritize uniformity ($\Pi^{d,c}$ compilations).

\begin{figure}
    \centering
    \includegraphics[width=0.75\columnwidth]{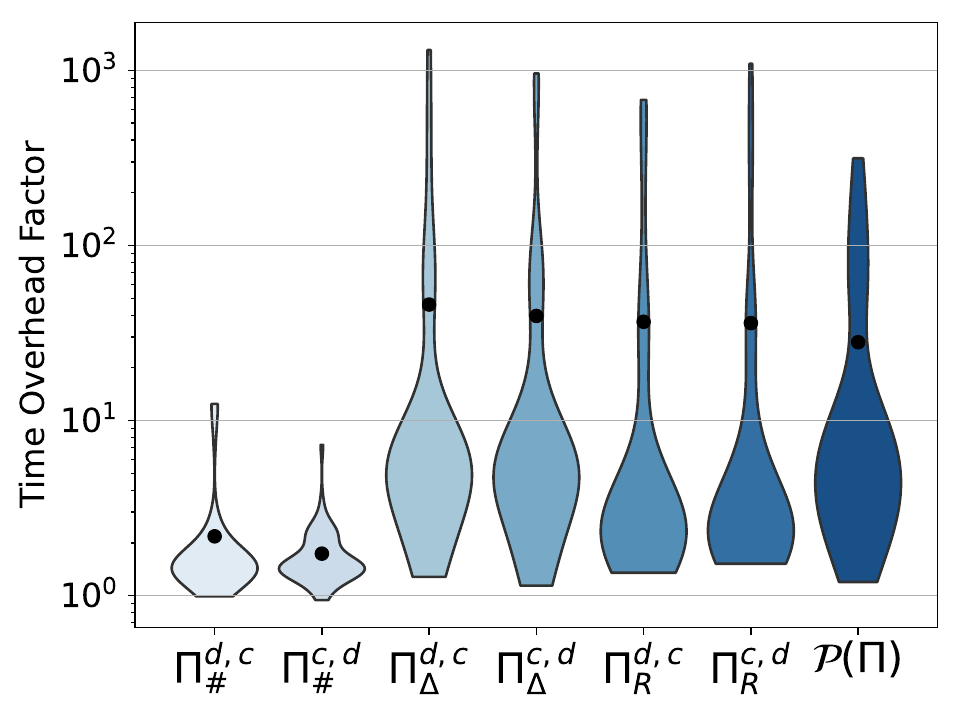}
    \caption{Distribution of the execution time overhead factor $\frac{T(X)}{T(\Pi)}$ for each approach $X$. Black dots represent the average.}
    \label{fig:execution_time_overhead}
\end{figure}

In order to further understand the overhead induced by our compilations, we compare the execution time $T$ needed by the planner to solve the new tasks $X$ versus the time needed to solve the standard planning task.
We refer to this as the \textit{time overhead factor}, and formally define it as $\frac{T(X)}{T(\Pi)}$.
This execution time includes both the time needed to translate (and ground) the task and the solving time.
To make the comparison fair, we only consider the $59$ problems that are commonly solved by all the approaches.
Figure~\ref{fig:execution_time_overhead} shows this analysis as a set of violinplots, which represent the distribution of these factors in log scale for each approach.
As we can see, minimizing the number of different action costs hardly introduces any overhead compared to solving the standard planning task, resulting into easier tasks in few problems.
On the other hand, the rest of the compilations are much closer in execution time to computing all the cost-optimal plans.

\subsubsection{Is Solving $\Pi$ Good Enough?}
\begin{figure}
    \centering
    \includegraphics[width=0.75\columnwidth]{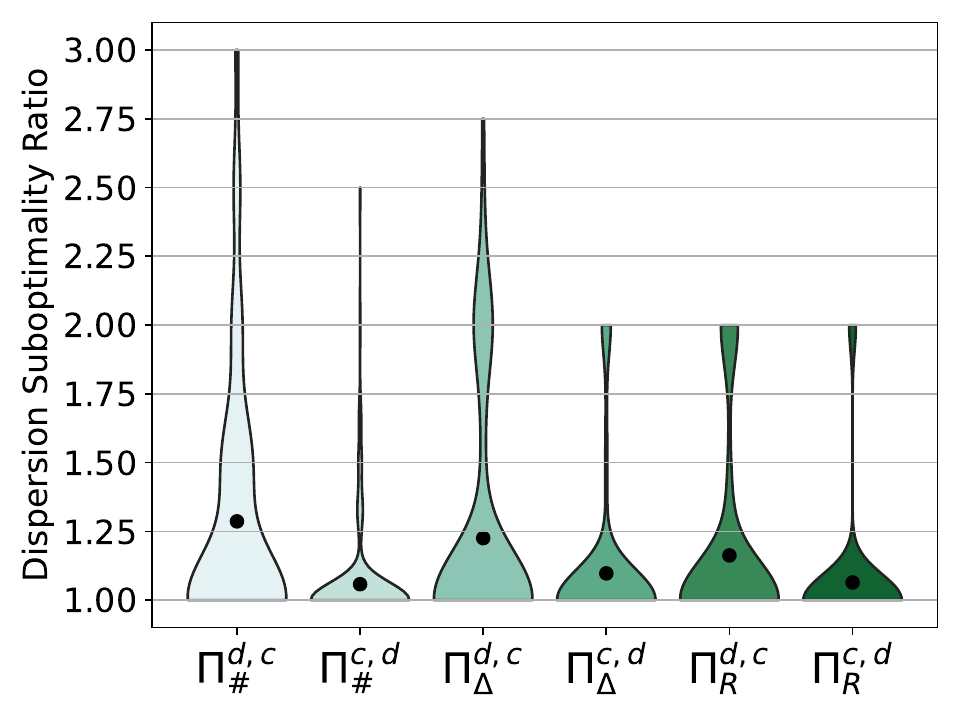}
    \caption{Distribution of the dispersion suboptimality ratio $\frac{d(\pi)}{d(\pi_{d})}$ for each dispersion metric $d$.}
    \label{fig:dispersion_suboptimality}
\end{figure}

With the coverage and execution time results at hand, one could arguably ask whether solving the standard planning task $\Pi$ is a good enough uniformity proxy.
In order to refute this hypothesis, we compare the dispersion metric $d$ induced by the plan obtained by the compilation that optimizes it, $\pi_d$, against the dispersion metric induced by the plan obtained by solving the standard planning task, $\pi$.
We refer to this as the \textit{dispersion suboptimality ratio}, and formally define it as $\frac{d(\pi)}{d(\pi_{d})}$.
Figure~\ref{fig:dispersion_suboptimality} shows this analysis as a set of violinplots, which represent the distribution of these ratios for each compilation.
Here we consider problems that were commonly solved by the given compilation and the standard planning task.
If we only focus on cost-optimal plans, i.e., only in the $\Pi^{c,d}$ compilations, the solution obtained by solving the standard planning task, $\pi$, also minimizes the given dispersion metric in most of the tasks.
On the other hand, if we focus on uniformest plans, i.e., only in the $\Pi^{d,c}$ compilations, we observe a larger suboptimality gap, meaning that $\pi$ is not as uniform as it could.

\begin{figure}
    \centering
    \includegraphics[width=0.8\columnwidth]{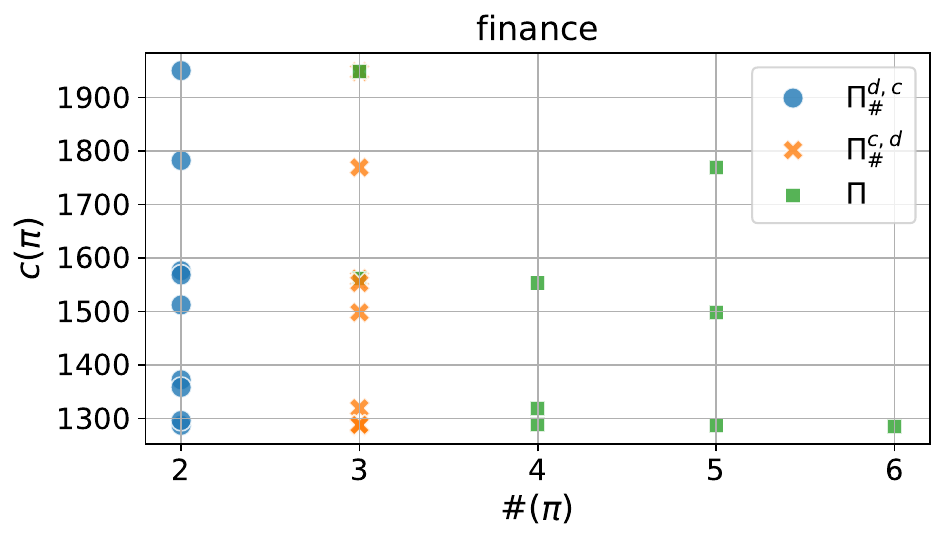}
    \caption{Cost ($y$) and number of different action costs ($x$) of the plans that solve $\Pi$, $\Pi_{\#}^{c,d}$, and $\Pi_{\#}^{d,c}$ tasks in \textsc{finance}.}
    \label{fig:finance}
\end{figure}

Although these results might initially suggest that solving the standard planning task provides uniform enough plans, that would be a conclusion drawn from a myopic analysis.
This is because most planning tasks have few cost-optimal plans involving very few different action costs. 
If we perform this analysis in tasks with plans that have different action costs, we can clearly observe that cost-optimal plans obtain dispersion metrics that are up to $3$ times suboptimal.
Figure~\ref{fig:finance} shows the cost and number of different action costs of the plans that solve $\Pi$, $\Pi_{\#}^{c,d}$, and $\Pi_{\#}^{d,c}$ tasks in \textsc{finance}.
As we can see, the cost-optimal plans (green squares) are the cheapest, but often involve more action costs.
On the other side of the spectrum we have $\Pi_{\#}^{d,c}$ (blue points), which computes uniformest plans, breaking ties in favor of less costly plans.
Solving the compilation allows us to get plans that, although not cost-optimal in some cases, only contain two different action costs.
In this domain, this means the plan is suggesting to always save the same amount of money every month (one action cost), plus the last zero-cost action that checks if the financial goal has been achieved.
$\Pi_{\#}^{c,d}$ (orange crosses) represent a good compromise between both approaches, as the plans generated are cost-optimal and require to use only three different action costs.

\subsubsection{How Some Dispersion Metrics Approximate Others. }
\begin{figure}
    \centering
    \includegraphics[width=0.9\columnwidth]{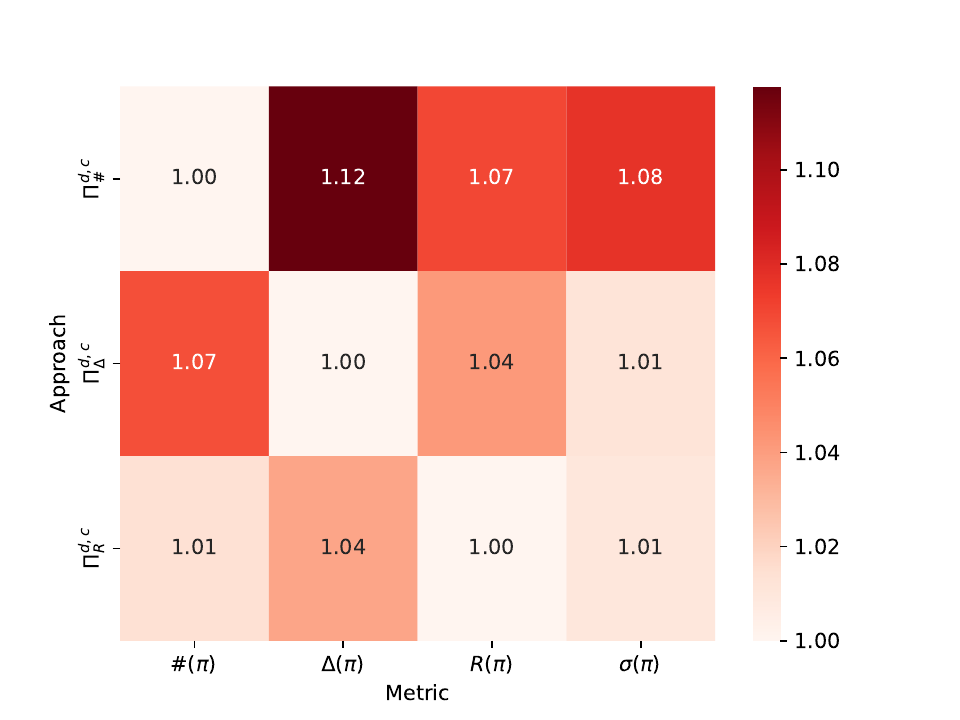}
    \caption{Dispersion suboptimality ratio for each combination of compilation (rows) and dispersion metric (columns). Lighter cells indicate better performance.}
    \label{fig:proxy}
\end{figure}

Finally, we want to understand whether optimizing for some dispersion metrics allow us to get good values in other metrics.
We perform this analysis with two objectives: (i) test if more scalable compilations such as $\Pi_{\#}$ also approximate Range or Delta; and (ii) test if some of these compilations also approximate well-established dispersion metrics for which we could not generate compilations such as standard deviation $\sigma$.
Figure~\ref{fig:proxy} shows the result of this analysis as a heatmap, where we have the compilations that generate uniformest plans in the rows, and the three different dispersion metrics plus standard deviation in the columns.
The values in the cells of the heatmap indicate the average dispersion suboptimality ratio for each combination of compilation and dispersion metric in the $90$ instances that were commonly solved by the three compilations.
For example, a value of $1.00$ in the $(\#(\pi), \Pi_{\#}^{d,c})$ cell indicates that, as expected, compilation $\Pi_{\#}^{d,c}$ was able to always find plans with optimal $\#(\pi)$ values.
On the other hand, the $1.12$ in the next column of the same row indicates that compilation $\Pi_{\#}^{d,c}$ returned plans with an average suboptimality of $1.12$ with respect to the Delta dispersion metric.
Since we cannot get the optimal standard deviation value for all the problems, we compute its dispersion suboptimality ratio by using the best $\sigma$ value found by any compilation in the denominator.

Cells with a lighter background indicate better (closer to the optimal) performance.
If we analyze each approach (full row), we can observe that all the compilations approximate quite well all the dispersion metrics, including standard deviation.
This makes sense, as all the compilations are in one way or another restricting the variety of the action costs used.
In particular, $\Pi_{\Delta}^{d,c}$ and $\Pi_{R}^{d,c}$, the most complex compilations, obtain the best results across all dispersion metrics.
On the other hand, the more scalable compilation $\Pi_{\#}^{d,c}$ shows only a slightly worse behavior, which makes it a good candidate as an affordable way of getting uniform plans.

\section{Conclusions and Future Work}
In this paper we have introduced a novel objective that might be relevant in many planning applications: that of finding plans whose actions have costs that are as uniform as possible.
We have adapted three different dispersion metrics to automated planning, and presented six different compilations to produce plans that lexicographically optimize sum of action costs and the given dispersion metric.
Experimental results across a large number of existing and novel planning tasks show that our methods are able to generate plans that optimally balance these two objectives.
While some of the compilations generate tasks that are hard to solve in practice, our compilations to generate plans with a low number of different action costs are able to scale on par with the standard planning task.
The plans generated by solving these more amenable compilations are also able to implicitly optimize more complex dispersion metrics, making them good candidates to compute uniform plans in practice.

In this work we solely focused on computing optimal solutions for uniformest cost-optimal planning problems.
In future work we would like to solve the compiled tasks using satisficing planners to study the trade-off between scalability and suboptimality.
Finally, we focused on actions cost uniformity, as cost usually encodes the effort of executing an action.
Extending our compilations to other definitions of uniformity such as preferring to use the same grounded (lifted) action is straightforward, and we would like to evaluate the resulting plans as part of future work.

\subsection*{Disclaimer}
This paper was prepared for informational purposes by
the Artificial Intelligence Research group of JPMorgan Chase \& Co. and its affiliates (``JP Morgan''),
and is not a product of the Research Department of JP Morgan.
JP Morgan makes no representation and warranty whatsoever and disclaims all liability,
for the completeness, accuracy or reliability of the information contained herein.
This document is not intended as investment research or investment advice, or a recommendation,
offer or solicitation for the purchase or sale of any security, financial instrument, financial product or service,
or to be used in any way for evaluating the merits of participating in any transaction,
and shall not constitute a solicitation under any jurisdiction or to any person,
if such solicitation under such jurisdiction or to such person would be unlawful.

\bibliography{aaai24}

\end{document}


\begin{table}[]
\setlength{\tabcolsep}{3pt}
    \centering
    \begin{tabular}{c|c|c|c|c|c|c|c|c}
        Domain & $\Pi$ & $\Pi_{\#}^{d,c}$ & $\Pi_{\#}^{c,d}$ & $\Pi_{\Delta}^{d,c}$ & $\Pi_{\Delta}^{c,d}$ & $\Pi_{R}^{d,c}$ & $\Pi_{R}^{c,d}$ & ${\cal P}(\Pi)$ \\ \hline
        agricola  & 0  & 0  & 0  & 0  & 0  & 0  & 0  & 0  \\ 
barman  & 4  & 4  & 4  & 4  & 4  & 4  & 4  & 3  \\
cybersec  & 9  & 1  & 1  & 0  & 0  & 0  & 0  & 0   \\ 
data-network  & 12  & 9  & 12  & 4  & 6  & 3  & 5  & 11   \\
elevators  & 18  & 15  & 16  & 4  & 5  & 4  & 6  & 6   \\ 
finance  & 19  & 9  & 18  & 0  & 0  & 0  & 0  & 19   \\
floortile  & 8  & 8  & 6  & 1  & 0  & 5  & 5  & 0   \\
gedit  & 15  & 15  & 15  & 15  & 15  & 15  & 15  & 15   \\ 
navigation  & 20  & 0  & 20  & 0  & 1  & 0  & 0  & 8   \\ 
openstacks  & 3  & 3  & 3  & 2  & 2  & 3  & 3  & 0   \\
organic-synthesis  & 17  & 15  & 15  & 10  & 10  & 6  & 6  & 10  \\ 
parcprinter & 14  & 17  & 16  & 0  & 9  & 0  & 12  & 3   \\ 
pegsol  & 19  & 18  & 18  & 17  & 17  & 18  & 18  & 17   \\ 
petri-net  & 9  & 9  & 9  & 9  & 9  & 9  & 9  & 0   \\
scanalyzer  & 13  & 8  & 13  & 4  & 10  & 11  & 11  & 5   \\
sokoban  & 20  & 20  & 20  & 20  & 20  & 20  & 20  & 11   \\ 
tetris  & 6  & 5  & 5  & 3  & 3  & 4  & 4  & 2   \\ 
transport  & 6  & 5  & 4  & 1  & 1  & 1  & 1  & 5   \\
woodworking  & 13  & 6  & 9  & 6  & 3  & 2  & 6  & 0  \\ \hline
\textsc{total} & $225$ & $167$ & $204$ & $100$ & $115$ & $105$ & $125$ & $115$
    \end{tabular}
    \caption{Number of problems solved by each compilation across the different domains.}
    \label{tab:coverage}
\end{table}